# Autistic Children's Mental Model of an Humanoid Robot

Cristina Gena, University of Turin, Turin, Italy, cristina.gena@unito.it

Claudio Mattutino, University of Turin, Turin, Italy, Claudio.mattutino@unito.it

Andrea Maieli, University of Turin, Turin, Italy, andrea.maieli@edu.unito.it

Elisabetta Miraglio, University of Turin, Turin, Italy, elisabetta.miraglio@unito.it

Giulia Ricciardiello, University of Turin, Turin, Italy, giulia.ricciardiello@unito.it

Rossana Damiano, University of Turin, Turin, Italy, rossana.damiano@unito.it

Alessandro Mazzei, University of Turin, Turin, Italy, alessandro.mazzei@unito.it

This position paper introduces the results of an initial card sorting experiment based on the reactions and questions of a group of children with autism working with a humanoid robot in a therapeutic laboratory on autonomy.

## 1  INTRODUCTION

Craik [1] described mental models as internal constructions of some aspect of the external world enabling predictions to be made, involving unconscious and conscious processes, where images and analogies are activated. Mental models are fundamental in HCI: according to Nielsen [4], a mental model is what the user believes about the system at hand, is based on beliefs, not facts, and individual users each have their own mental model. System designers should construct conceptual models of the systems inspiring the creation of a correct user's mental model.

Researchers, working on believable agents and emotional robots, have used anthropomorphism in their design to inspire a user's mental model. However, in order to test their design, they have also developed reliable and valid measures of people's responses to robots, social agents, and robotic assistants [3].

For instance, the Godspeed scale [1] was developed as a way to measure human and robotic interaction, and it has become a widely used standard. There are five central dimensions to the Godspeed scale: i) *anthropomorphism*, or the extent to which a robot appears humanlike versus machinelike; ii) *animacy*, or how lifelike a robot seems; iii) *likeability*, or how friendly a robot seems; iv) p*erceived intelligence of the robot*; and v) *perceived safety*, or emotional state/anxiety of the perceiver. [3] drawing from the Godspeed Scale and from the psychological literature on social perception, develop an 18-item scale (The Robotic Social Attribute Scale; RoSAS) to measure people's judgments of the social attributes of robots. Factor analyses reveal three underlying scale dimensions: warmth (feeling, happy, organic, compassionate, social, and emotional), competence (the intelligence or ability of the robot), and discomfort (knowledgeable, interactive, responsive, capable, competent, and reliable).

Inspired by the Godspeed scale's dimensions as a measure of anthropomorphism, in this position paper we report the first results of an inductive analysis on the questions posed to and on the comments/reaction to the Pepper humanoid robot by a group of four children with autism aged from 11 to 13, involved in a laboratory on autonomy.

## 2  USE CASE AND ANALYSIS

The Sugar, Salt & Pepper - Humanoid robotics for autism - research project focuses on the use of the Pepper robot in a therapeutic laboratory on autonomy that aims to promote functional acquisitions in highly functioning (Asperger) children with autism.

During the laboratory, started at the end of February 2021 in an apartment of the Paideia Foundation, we wanted to test the exchanges and interactions of children in rehabilitation contexts with the robot helping the operators. Another goal of the project is to provide young participants with a space for increasing their skills of mutual communication and socialization and strengthen the acquisition of strategies related to daily activities, such as the preparation of a snack or the management of homework, with the help of Pepper, configured as a highly motivating and engaging tool.

The weekly meetings lasted one hour and are led by a therapist. Each meeting had this structure: welcome in the apartment managed by the robot and the therapists; dialogue session with the robot on a predetermined topic (e.g. music, video games, etc.); moment of snack preparation; moment of post-snack dialogue and final feedback. After 5 meetings, we started analyzing the questions that children spontaneously ask the robot and the comments they make aloud.

Four analysts separately analyzed the transcription of the meetings' notes, created labels that summarized the prevailing concepts, and then we did a cardsorting for converging on the same categories from which to extrapolate the mental images that the children have made about the robot. We will use these categories as points of view to better build the robot's personality and character and later we will use them as semantic classes to establish the types of dialogue that the robot should best provide. The emerging categories have been the following ones.

**Personal life**. The children wanted to know what Pepper eats and drinks, which are its parents, which is its temperature, etc. It seems that they tend to attribute it to biological and social life.

**Preferences and interests**. Related to their expectations on its social life, children asked the robot questions such as which kind of music it likes, and if it likes going to school.

**Intelligence and knowledge**. Children showed great expectations about the intellectual capabilities and knowledge of the robot, witnessed by questions on ability to make difficult calculations, or on its knowledge about particular topics, almost encyclopedic, and sometimes surprising such as Shintoism, etc.

**Skills and actions**. As well as kids having expectations on the exceptional abilities of calculation and knowledge the robot should have, they also have great expectations on the exceptional things it can do, such as speaking foreign languages perceived as difficult such as Japanese, playing the piano, making war, etc.

**Intentionality and purposes**. Children attribute to the robot also intents and purposes, often linked to what the robot has told them in the past (Robot promises us things he has to keep). On the one hand, they would like to know what are its contextual goals related to its presence at the laboratory, on the other hand they are curious about its long-term goals and intentions: does it want to conquer the world?

**Physical aspect and hardware**. Some children have asked it about its body features (do you have prehensile hands like iCub robot?) or have made comments like what is its blinking good for?, or it should have sensors to be able to move independently, its fan may fatigues it. Others have asked about its motherboard.

## 3 CONCLUSION

From this initial analysis, we realized that the robot personality and its character should be better defined and detailed in order to satisfy the curiosity and the expectations of children, and to make it more credible and engaging, especially for an educational context [5]. At the moment, the robot often does not know how to answer the questions or does not understand them, and this generates friction and nervousness in children. So, we realized that we need to work more on its dialogue strategies and on enriching its knowledge. At the dialogue level, giving the robot a richer personality that it can convey through language, would constrain the expectations of the children to a more tractable set of topics, and create justifications for its inadequacies.

As future work, we are working on the adaptive mechanisms with which Pepper will be enriched. The robot will show more and more intelligence and reasoning skills, which will allow it to adapt to the user's needs and customize the interaction in both adaptive and adaptable form, as described in [6]. Thus, the robot will become an adaptive robot with respect to the user, able to adapt its behavior based on the characteristics of the subject with whom it is interacting and to customize the interaction in an adaptive form, that is, configurable by the staff expert in therapy. However, from our initial investigation we believe that the interaction should be "robot-guided" in order to give smarter answers to the (big variety) of questions collected.

## 4 ACKNOWLEDGMENTS

The multidisciplinary project is carried out by the University of Turin, the Paideia Foundation, Intesa Sanpaolo Innovation Center and Jumple, and was funded by Banca Intesa Sanpaolo, Banca dei Territori Division.